\documentclass[letterpaper]{article} % DO NOT CHANGE THIS
\usepackage{aaai25}  % DO NOT CHANGE THIS
\usepackage{times}  % DO NOT CHANGE THIS
\usepackage{helvet}  % DO NOT CHANGE THIS
\usepackage{courier}  % DO NOT CHANGE THIS
\usepackage[hyphens]{url}  % DO NOT CHANGE THIS
\usepackage{graphicx} % DO NOT CHANGE THIS
\usepackage{natbib}  % DO NOT CHANGE THIS AND DO NOT ADD ANY OPTIONS TO IT
\usepackage{caption} % DO NOT CHANGE THIS AND DO NOT ADD ANY OPTIONS TO IT
\usepackage[linesnumbered,ruled]{algorithm2e}
\usepackage{multirow}
\usepackage{pifont}
\usepackage[dvipsnames]{xcolor}
\usepackage{colortbl}
\usepackage{xspace}
\newcommand{\etal}{\textit{et al.}\xspace}
\newcommand{\eg}{\textit{e.g.}\xspace}
\newcommand{\ie}{\textit{i.e.}\xspace}
\newcommand{\wrt}{w.r.t.\xspace}

\usepackage{enumitem}

\usepackage{amsmath}
\usepackage{booktabs}
\usepackage{amssymb}
\usepackage{subcaption}

\usepackage{bm}

\usepackage[capitalize]{cleveref}
\crefname{section}{Sec.}{Secs.}
\Crefname{section}{Section}{Sections}
\crefname{table}{Tab.}{Tabs.}
\Crefname{table}{Table}{Tables}
\crefname{algorithm}{Alg.}{Algs.}
\Crefname{algorithm}{Algorithm}{Algorithms}

\title{Enhancing Contrastive Learning Inspired by the Philosophy of\\ ``The Blind Men and the Elephant''}

\author {
    Yudong Zhang\textsuperscript{\rm 1,\rm 2},
    Ruobing Xie\textsuperscript{\rm 2,}\equalcontrib,
    Jiansheng Chen\textsuperscript{\rm 3,}\equalcontrib,
    Xingwu Sun\textsuperscript{\rm 2,\rm 4},
    Zhanhui Kang\textsuperscript{\rm 2},
    Yu Wang\textsuperscript{\rm 1,}\equalcontrib
}
\affiliations {
    \textsuperscript{\rm 1}Tsinghua University\\
    \textsuperscript{\rm 2}Tencent\\
    \textsuperscript{\rm 3}University of Science and Technology Beijing\\
    \textsuperscript{\rm 4}University of Macau\\
    zhangyd16@mails.tsinghua.edu.cn, xrbsnowing@163.com, jschen@ustb.edu.cn, sunxingwu01@gmail.com, kegokang@tencent.com, yu-wang@mail.tsinghua.edu.cn
}

\usepackage{bibentry}
\begin{document}

\maketitle

\begin{abstract}
Contrastive learning is a prevalent technique in self-supervised vision representation learning, typically generating positive pairs by applying two data augmentations to the same image. Designing effective data augmentation strategies is crucial for the success of contrastive learning. Inspired by the story of the blind men and the elephant, we introduce JointCrop and JointBlur. These methods generate more challenging positive pairs by leveraging the joint distribution of the two augmentation parameters, thereby enabling contrastive learning to acquire more effective feature representations. 
To the best of our knowledge, this is the first effort to explicitly incorporate the joint distribution of two data augmentation parameters into contrastive learning. As a plug-and-play framework without additional computational overhead, JointCrop and JointBlur enhance the performance of SimCLR, BYOL, MoCo v1, MoCo v2, MoCo v3, SimSiam, and Dino baselines with notable improvements.
\end{abstract}

\begin{links}
    \link{Code}{https://github.com/btzyd/JointCrop}
    % \link{Extended version}{https://github.com/btzyd/JointCrop/appendix.pdf}
\end{links}

\begin{figure}[!t]
    \centering
    \begin{subfigure}{0.49\linewidth}
        \includegraphics[width=\linewidth]{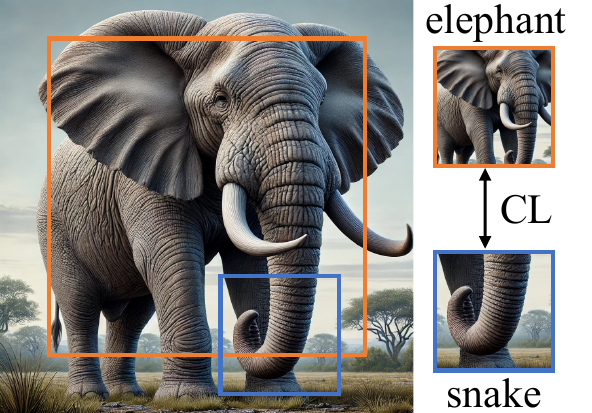}
        \caption{JointCrop generates challenging positive pairs.}
        \label{fig:fig1a}
    \end{subfigure}
    \begin{subfigure}{0.49\linewidth}
        \includegraphics[width=\linewidth]{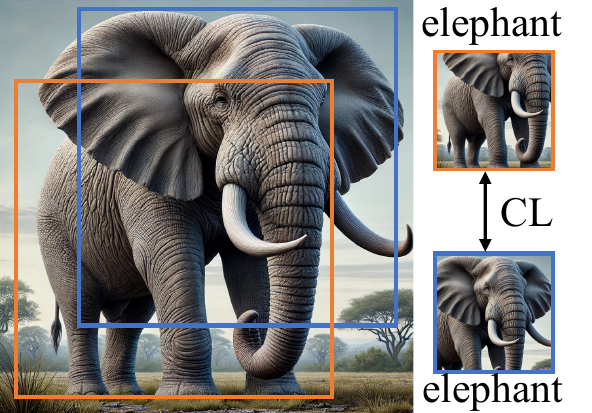}
        \caption{RandomCrop generates trivial positive pairs.}
        \label{fig:fig1b}
    \end{subfigure}
    \hfill
    \begin{subfigure}{0.49\linewidth}
        \includegraphics[width=\linewidth]{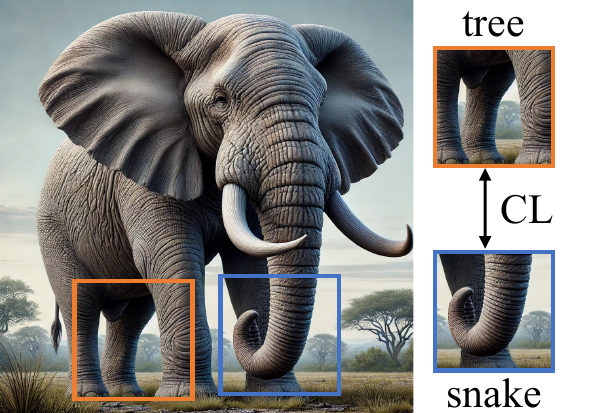}
        \caption{RandomCrop generates incomprehensive positive pairs.}
        \label{fig:fig1c}
    \end{subfigure}
    \begin{subfigure}{0.49\linewidth}
        \includegraphics[width=\linewidth]{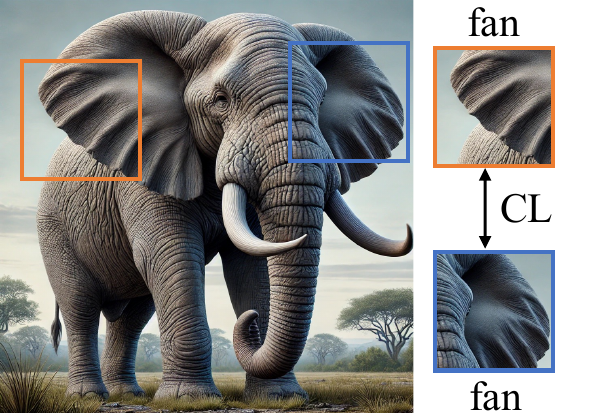}
        \caption{RandomCrop generates trivial and incomprehensive pairs.}
        \label{fig:fig1d}
    \end{subfigure}
    \hfill
    \begin{subfigure}{0.49\linewidth}
        \includegraphics[width=\linewidth]{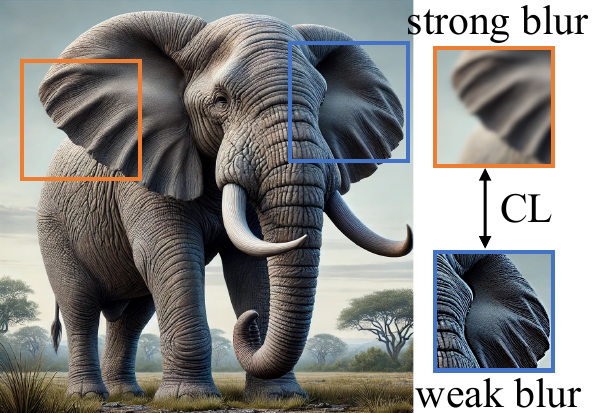}
        \caption{JointBlur generates challenging positive pairs.}
        \label{fig:fig1e}
    \end{subfigure}
    \begin{subfigure}{0.49\linewidth}
        \includegraphics[width=\linewidth]{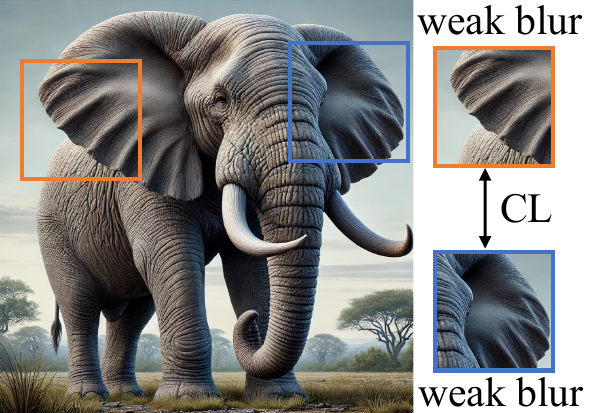}
        \caption{GaussianBlur generates trivial positive pairs.}
        \label{fig:fig1f}
    \end{subfigure}
    \caption{The motivation of our paper. We use the philosophy of \textit{the blind men and the elephant} to analyze contrastive learning between positive sample pairs.}
    \label{fig:figure1}
\end{figure}

% Uncomment the following to link to your code, datasets, an extended version or similar.
%

\section{Introduction}
\label{sec:introduction}

Self-supervised learning (SSL) \cite{swav,smog} has garnered significant attention in recent years as obtaining large amounts of labeled data is expensive. Contrastive learning, a widely-utilized SSL method, can even outperform supervised learning on tasks such as image classification, object detection, and semantic segmentation \cite{mocov3,dino}.

Contrastive learning (CL) produces self-supervised signals through pretext tasks and utilizes these signals to train the encoder. A common pretext task is instance discrimination (ID) \cite{inst_dise}, which involves a classification problem at the instance level. 
A pair of positive views is generated by applying two independently distributed data augmentations to a single image, aiming to maximize the similarity of their representations. 
ID is widely utilized in popular CL methods such as SimCLR \cite{simclr}, MoCo \cite{mocov1, mocov2}, BYOL \cite{byol}, SimSiam \cite{simsiam}, SWAV \cite{swav}, and DINO \cite{dino}.

The design of positive pairs is crucial for CL. Previous studies have employed various data augmentations to generate positive pairs, including color distortion, puzzle transformations, and adversarial attacks \cite{jiang2020robust,kim2020adversarial,ho2020contrastive}. 
The InfoMin \cite{infomin} approach reduces the mutual information between positive pairs while preserving task-relevant information to enhance transfer performance. Zhu \etal \cite{zhu2021improving} obtain additional positive views via positive extrapolation. ContrastiveCrop \cite{ccrop} crops images based on heatmaps to ensure the presence of objects in the cropped views. 

Let's begin by revisiting the time-honored story of \textit{the Blind Men and the Elephant} \cite{saxe1884poems}. In this tale, six blind men each encounter a different part of an elephant. The man who touches the side believes the elephant is a wall, while those who feel the tusks, trunk, knees, ears, and tail conclude that it resembles a spear, snake, tree, fan, and rope, respectively. Despite their stubbornness and arguments, none have seen the entire elephant. In the context of this story, Contrastive learning enhances the understanding of elephants by decreasing the feature distance between positive pairs, much like how people deepen their understanding of elephants through argumentation.

From the standpoint of difficulty of positive pairs, as Zhu suggests, more challenging positive pairs can help contrastive learning acquire better representations \cite{zhu2021improving}. Establishing connections between the whole and the parts of an elephant is more challenging than dealing with incomprehensive or trivial cases. Let's use the elephant example to illustrate this: (1) In \cref{fig:fig1b}, both blind men perceive the whole elephant. Although they both have a global perception, the task is trivial and not sufficiently challenging. (2) In \cref{fig:fig1c}, the two blind men perceive an elephant's leg and trunk, mistaking them for a tree and a snake, respectively. Since neither has global information, their interaction results in an incomprehensive and one-sided perception of the elephant, leaving them unaware that it is an elephant at all. (3) In \cref{fig:fig1d}, both blind men perceive the elephant's ears and mistake them for fans. This interaction is neither constructive nor sufficiently challenging, and it fails to lead to a comprehensive understanding of the elephant. (4) In \cref{fig:fig1a}, one blind person perceives the whole elephant while the other focuses on the trunk. Through their interaction, they both develop a deeper understanding of the elephant as a whole, as well as the specific details of the trunk. This task is both non-trivial and challenging. 

In summary, we argue that CL can benefit from forming more challenging connections between global and local information. However, we have observed that existing CL methods often fall short in generating sufficiently diverse samples, resulting in many samples that are inherently similar or lack comprehensiveness. By analyzing the distribution of area ratios between pairs of positive samples in current CL methods, we found that 80\% of RandomCrop pairs have area ratios within 1:2. This leads to a large number of cases similar to those illustrated in \cref{fig:fig1b,fig:fig1c,fig:fig1d}.

In addition to RandomCrop, we can also consider the data augmentation technique GaussianBlur, which is commonly used in contrastive learning. This technique is akin to putting myopic glasses on the observer; despite the blur, it does not hinder the observer's ability to recognize the elephant or its parts. In \cref{fig:fig1e}, the combination of normal observations (weak blur) and myopic observations (strong blur) results in challenging samples. Conversely, the positive pairs in \cref{fig:fig1f}, both employing weak blur, are more trivial.

We refer to the methods we used to generate more challenging positive pairs as JointCrop (\cref{fig:fig1a}) and JointBlur (\cref{fig:fig1e}). Previous studies have generated more challenging positive pairs by employing stronger or more diverse data augmentations; however, the two data augmentations applied to positive sample pairs remain independent. In contrast, our proposed methods, JointCrop and JointBlur, \textbf{first establish a specific relationship between the positive pairs and then determine the parameters used to generate the two positive samples}. In other words, while previous studies assume that the joint distribution of the two data augmentations is simply the product of their marginal distributions, our approach ensures that the two data augmentations of the positive pairs are interdependent. 

Specifically, we intentionally manage a certain metric that effectively measures the difficulty level between positive pairs and use this metric to control the parameters of the two data augmentations. This metric indirectly induces a correlation between the two augmentation parameters, effectively allowing us to control their joint distribution. Consequently, we refer to our methods as JointCrop and JointBlur, which build upon RandomCrop and GaussianBlur, respectively. 

Our plug-and-play JointCrop and JointBlur methods are agnostic to CL methods and do not require considerations such as the use of negative samples. Additionally, they incur negligible additional computational overhead during training. As a ``free lunch'', our JointCrop and JointBlur offer non-trivial improvements over the SimCLR, BYOL, MoCo v1, MoCo v2, MoCo v3, SimSiam, and Dino baselines. Furthermore, our JointCrop and JointBlur can also be used in conjunction with existing techniques for enhancing contrastive learning, such as Multi-Crop and ContrastiveCrop, to further improve the performance of contrastive learning.

The main contributions of this study are summarized as follows: (1) To the best of our knowledge, this is the first study to explicitly introduce the correlation of data augmentations between positive pairs in contrastive learning. (2) We introduce JointCrop and JointBlur, which generate more challenging samples by controlling the distribution of area ratios and GaussianBlur kernels between positive pairs. Additionally, we abstracted a unified framework, JointAugmentation, from both methods, paving the way for this concept to be applied to a broader range of data augmentations. (3) As a ``free lunch'', our plug-and-play approach incurs no additional computational cost and enhances baselines across various datasets and popular contrastive learning methods.

\section{Related Work}
\label{sec:relatedworks}

%------------------------------------------------------------------------
\subsection{Contrastive Learning}
\label{sec:contrastive-learning}

Contrastive learning is a self-supervised learning approach pretrained by pretext tasks with unlabeled data. 
Previous studies have designed challenging augmented samples to supervise the encoder in learning better feature representations \cite{bachman2019learning,misra2020self,inst_dise,ye2019unsupervised}.
CL has achieved strong performance in the case of learning feature representations without labels, and the pretrained models are easy to transfer to downstream tasks such as classification, object detection, and instance segmentation. Contrastive learning achieves strong performance across many tasks \cite{liang2024factorized, feng2023maskcon, chanchani2023composition, xiao2024simple, sarto2023positive, li2023contrastive, wu2024voco, park2024self}.

CL can be categorized into two types based on the explicit use of negative samples. 
The contrastive learning methods that utilize both positive and negative samples include SimCLR \cite{simclr} and MoCo \cite{mocov1}. The core idea of these methods is maximizing the similarity between positive pairs, while minimizing the similarity between non-positive pairs.
The CL methods using only positive pairs, such as BYOL \cite{byol} and SimSiam \cite{simsiam}, uses siamese network structures and feeds pairs of positive views into them. 
Special designs, such as stop-grad \cite{simsiam}, momentum encoder \cite{mocov1}, and a predictor, are necessary to prevent model collapse in the absence of negative samples. 

Some studies categorize BYOL and SimSiam as non-contrastive methods. However, to explore the generalizability of our methods, we consider these representational learning approaches as CL methods, as they work by reducing the feature distance between positive pairs.

%------------------------------------------------------------------------

\subsection{Design of Positive Pairs}
\label{sec:designofpositivesamples}

Regardless of whether negative samples are used, the design of generating pairs of positive views is critical to CL.
A popular method to generate a pair of two positive views is applying two data augmentations on a specific image. Several studies have explored the design of positive pairs. SimCLR \cite{simclr} examines the effectiveness of different combinations of multiple augmentation method and finds that the most useful augmentation methods are Crop and Color. Some studies \cite{jiang2020robust,kim2020adversarial,ho2020contrastive} introduce adversarial attacks and use adversarial examples as positive or negative samples. Several methods have been proposed to craft positive pairs for contrastive learning. For example, ContrastiveCrop \cite{ccrop} leverages model heatmaps to guide the cropping region, thereby reducing the likelihood of excluding objects from the cropped area. Similarly, MultiCrop \cite{swav} replaces a single high-resolution sample with multiple low-resolution crops, thereby improving contrastive learning performance without a substantial increase in computational cost. InfoMin \cite{infomin} finds the sweet spot of mutual information between views and generates positive pairs. However, most of these methods do not explicitly address the questions of whether the two augmentations should be correlated and how they should be correlated. In this work, we propose JointCrop and JointBlur, which introduce the correlation between the augmentation parameters of positive pairs and consider their joint distribution, leading to more challenging views for CL without incurring additional overhead.

\section{Method}
\label{sec:method}

\subsection{Preliminaries}
\label{sec:preliminary}
We briefly review the pipeline of CL. For an input image $I$, CL generates a pair of positive samples by applying the data augmentation $\mathcal{T}$ twice, as shown in \cref{eq:ramdom_positive_paris}, where the cumulative distribution function $F(\bm{t})$ is the distribution used to sample the augmentation parameters $\bm{t}$. The views $v_{I,1}$ and $v_{I,2}$ form a pair of positive views, while other views from the images other than $I$ are considered negative samples.

\begin{equation}
    \begin{aligned}
        v_{I,1}=\mathcal{T}\left(I;{\bm{t_1}}\right)&, v_{I,2}=\mathcal{T}\left(I;{\bm{t_2}}\right)\\
        {\bm{t_1}}\sim{F(\bm{t_1})}&, {\bm{t_2}}\sim{F(\bm{t_2})}
    \end{aligned}
    \label{eq:ramdom_positive_paris}
\end{equation}

In CL methods without negative samples, such as BYOL \cite{byol} and SimSiam \cite{simsiam}, the representations of $v_{I,1}$ and $v_{I,2}$ are expected to be sufficiently similar. While in SimCLR \cite{simclr} and MoCo \cite{mocov1}, which are CL methods with negative samples, the representations of $v_{I,1}$ and $v_{I,2}$ are expected to be sufficiently close, and their representations are expected to be as distant as possible from the other negative views.

\subsection{JointCrop}
\label{sec:jointcrop}

To introduce our JointCrop method, we first review the RandomCrop pipeline. Initially, the area ${s}$ is randomly selected within a specified range, defined by ${s}\sim\mathcal{U}[s_\text{min},s_\text{max}]$. Next, the aspect ratio ${r}$ is also randomly selected. Given the area ${s}$ and aspect ratio ${r}$, we can uniquely determine the width ${w}=\sqrt{{s}\times{r}}$ and height ${h}=\sqrt{{s}/{r}}$. Following this, crop positions ${i}$ and ${j}$ are selected using ${i}\sim\mathcal{U}[0,\text{W}-{w}]$ and ${j}\sim\mathcal{U}[0,\text{H}-{h}]$, where W and H are the image's width and height, respectively. By repeatedly applying this sampling procedure twice to an image $I$, we obtain a positive pairs $v_{I,1}=\text{Crop}\left(I;{\bm{t_1}}=\left({i_1},{j_1},{h_1},{w_1}\right)\right)$ and $v_{I,2}=\text{Crop}\left(I;{\bm{t_2}}=\left({i_2},{j_2},{h_2},{w_2}\right)\right)$. Based on our previous analysis in \cref{fig:figure1}, the area ratio ${s_r} = \frac{{s_2}}{{s_1}} = \frac{{h_2}{w_2}}{{h_1}{w_1}}$ between positive pairs can significantly impact contrastive learning performance. To investigate this, we define a quantitative measure of the difficulty of data augmentation, termed Statistical Difficulty Factor (SDF). SDF measures the cosine similarity between all positive pairs generated by a data augmentation method $\mathcal{T}$ across the entire dataset $\mathcal{D}$, using an already trained contrastive learning SimSiam model $f$.

\begin{equation}
 \text{SDF}(\mathcal{T})=\mathbb{E}_{I\in\mathcal{D}}\left[\cos\left(f\left(v_{I,1}\right),f\left(v_{I,2}\right)\right)\right]
  \label{eq:sdf}
\end{equation}

\begin{figure}[h]
    \centering
    \includegraphics[width=\linewidth]{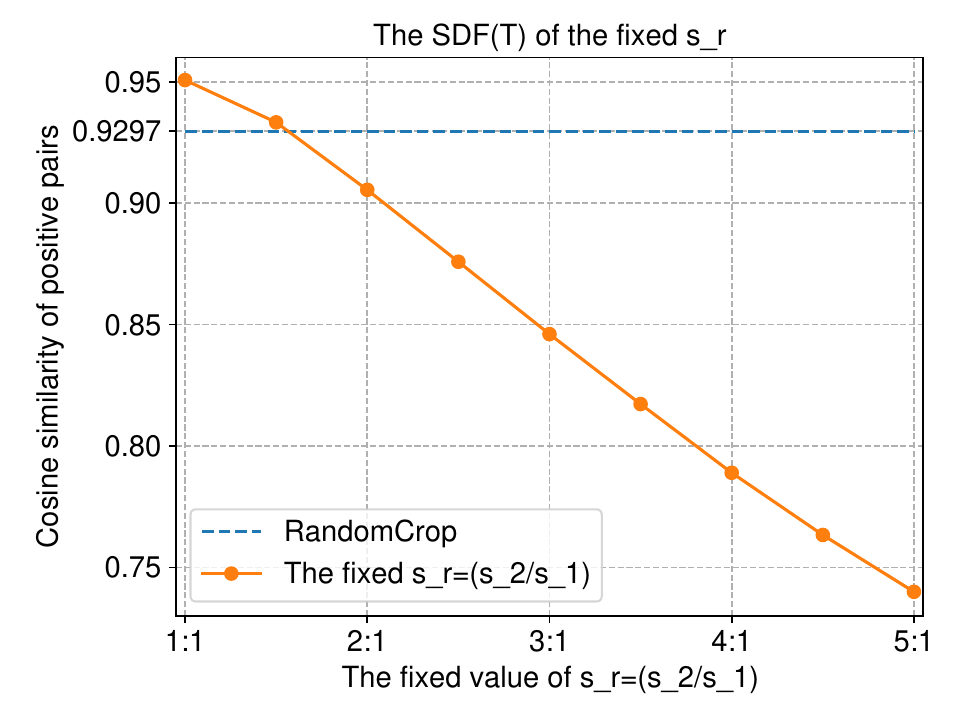}
    \caption{The statistical difficulty between the positive pairs generated by different fixed area ratios $s_r=s_2/s_1$.}
    \label{fig:fixed_s1:s2}
\end{figure}

We measured the values of SDF for RandomCrop and fixed area ratios ranging from 1:1 to 1:5. \Cref{fig:fixed_s1:s2} confirms our analysis: positive pairs become more challenging as the area ratio increases, while RandomCrop often results in trivial positive pairs. Therefore, we can use the area ratio as a means to control the difficulty level between positive pairs.

\begin{figure}[h]
    \centering    
    \includegraphics[width=\linewidth]{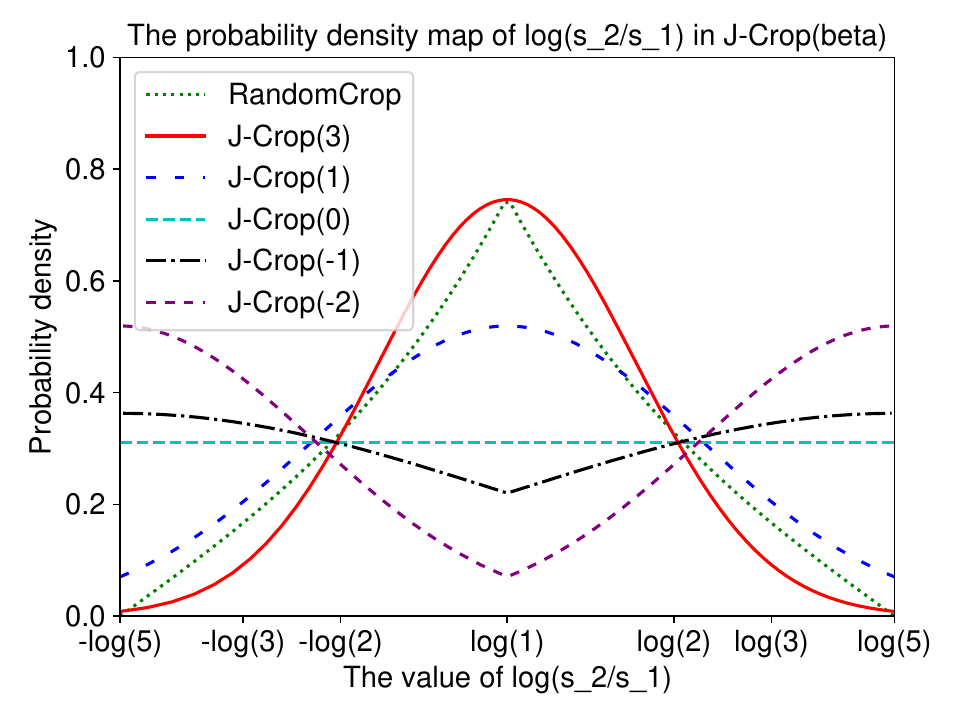}
    \caption{The probability density map of JointCrop, which controls the area ratios of positive pairs obeying a series of distributions JC$(\beta)$ controlled by $\beta$. The smaller $\beta$ leads to the higher probability that the ratios are far from 1.}
    \label{fig:pdf_jointcrop}
\end{figure}

To further investigate the distribution of area ratios in the RandomCrop, we aim to find the distribution of ${s_r} = {{s_2}}/{s_1}$ in RandomCrop. This can be formulated as a mathematical problem: given ${s_1}\sim\mathcal{U}[s_\text{min}, s_\text{max}]$ and ${s_2}\sim\mathcal{U}[s_\text{min}, s_\text{max}]$ (with a typical setup in RandomCrop being $s_\text{min}=0.2$ and $s_\text{max}=1$), we seek the distribution of ${s_r} = {s_2}/{s_1}$, the derivation of which is shown in \cref{sec:jifen}. Since ${s_r}$ takes values in the range $[0.2, 5]$, and to create symmetry in its probability density map about ${s_r}=1$, we plot the probability density map of $\log{s_r}$ as the green dotted line in \cref{fig:pdf_jointcrop}. We find that RandomCrop is more likely to produce samples with similar areas; for example, the probability that the area ratios between positive pairs exceed 2:1 is only $18.75\%$.

Larger area ratios between pairs of two views indicate more challenging samples, which can facilitate learning better representations. However, RandomCrop does not provide a sufficient number of samples with large area ratios. The goal of JointCrop is to make the probability density map of $\log{s_r}$ ``shorter'' and ``fatter'' compared to that of RandomCrop, as depicted by the green dotted line in \cref{fig:pdf_jointcrop}. A straightforward approach is to control ${s_r}$ to be as far from 1 as possible before sampling $s_1$ and $s_2$. To achieve this, our proposed JointCrop method controls the distribution of ${s_r} = \frac{{s_2}}{{s_1}}$ instead of sampling ${s_1}$ and ${s_2}$ from independent uniform distributions, as in RandomCrop. In other words, JointCrop manages the joint distribution of ${s_1}$ and ${s_2}$.

Specifically, we define a series of distributions, JC$(\beta)$. For each generation of two positive samples from a single image, JointCrop first samples $\log{s_r}\sim \text{JC}(\beta)$. Then, it samples ${s_1}\sim\mathcal{U}\left[\max\left(s_\text{min},\frac{s_\text{min}}{{s_r}}\right),\min\left(\frac{s_\text{max}}{{s_r}},s_\text{max}\right)\right]$. The value of ${s_2}$ is then directly calculated as ${s_2}={s_1}\times{s_r}$.
Because RandomCrop limits the minimum crop scale to $s_\text{min}$ and the maximum crop scale to $s_\text{max}$, we ensure that ${s_1},{s_2}\in [s_\text{min},s_\text{max}]$ in JointCrop by controlling the upper and lower bounds of the distribution of ${s_1}$. If this sampling yields ${s_r}>1$, only the first term of $\max(\cdot, \cdot)$ and $\min(\cdot, \cdot)$ is effective, otherwise the second term is effective.

We define JC$(\beta)$ in terms of a series of variants of the truncated Gaussian distribution, as shown in \cref{alg:jc,fig:pdf_jointcrop}. The truncated Gaussian distribution $\mathcal{N}_T(\mu,\sigma,p,q)$ has four parameters, $\mu$ and $\sigma$ denote the mean and standard deviation, respectively, while $p$ and $q$ represent the truncated minimum and maximum values.
Since $\log{s_r}$ is symmetric about $0$ and takes values in the range $[-s_b, s_b]$, where $s_b=\log\frac{s_\text{max}}{s_\text{min}}$, we define JC$(\beta>0) = \mathcal{N}_T(0,\frac{1}{\beta}s_b,-s_b,s_b)$. We generalize the definition of JC$(\beta)$ to $\beta = 0$, meaning $\sigma=\frac{1}{\beta}s_b\to\infty$, and in this case, JC$(0)$ approaches a uniform distribution $\mathcal{U}[-s_b,s_b]$. To create more challenging samples than JC$(0)$, we define JC$(\beta<0)$ by flipping the left and right halves of JC$(\lvert\beta\rvert)$ about $-\frac{s_b}{2}$ and $\frac{s_b}{2}$, respectively. The probability density map of JC$(\beta)$ is depicted in \cref{fig:pdf_jointcrop}. As $\beta$ decreases, the distribution of JC$(\beta)$ becomes ``shorter'' and ``fatter'', which implies the generation of more challenging positive pairs.

\begin{figure}[h]
    \centering
    \includegraphics[width=\linewidth]{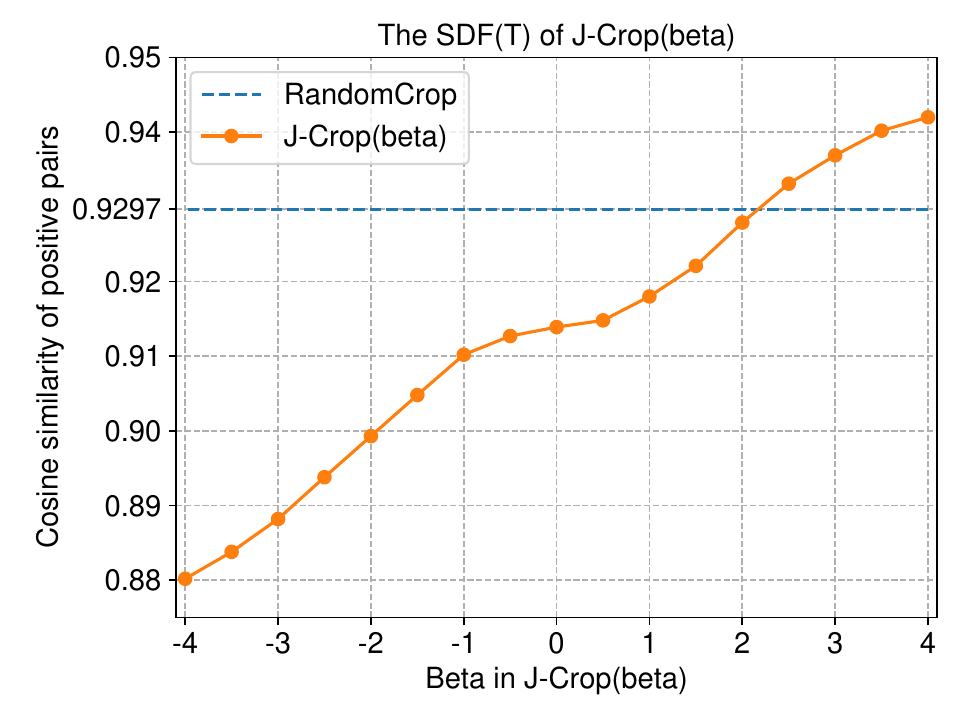}
    \caption{The SDF$(\mathcal{T})$ between the positive pairs generated by J-Crop$(\beta)$ is measured using the already trained SimSiam encoder on the whole ImageNet-1K training dataset.}
    \label{fig:simsiam_cos}
  
\end{figure}

The process of generating positive pairs using the area ratios $\log s_r\sim\text{JC}(\beta)$ is referred to as J-Crop$(\beta)$, with the steps detailed in \cref{alg:jc} in \cref{sec:algorithm}. We feed the positive pairs generated by J-Crop$(\beta)$ into a pre-trained SimSiam encoder and measure $\text{SDF}(\mathcal{T})$ of $\text{J-Crop}(\beta)$. The results, shown in \cref{fig:simsiam_cos}, indicate that smaller $\beta$ values result in more challenging samples compared to RandomCrop.

\begin{figure}[h]
    \includegraphics[width=\linewidth]{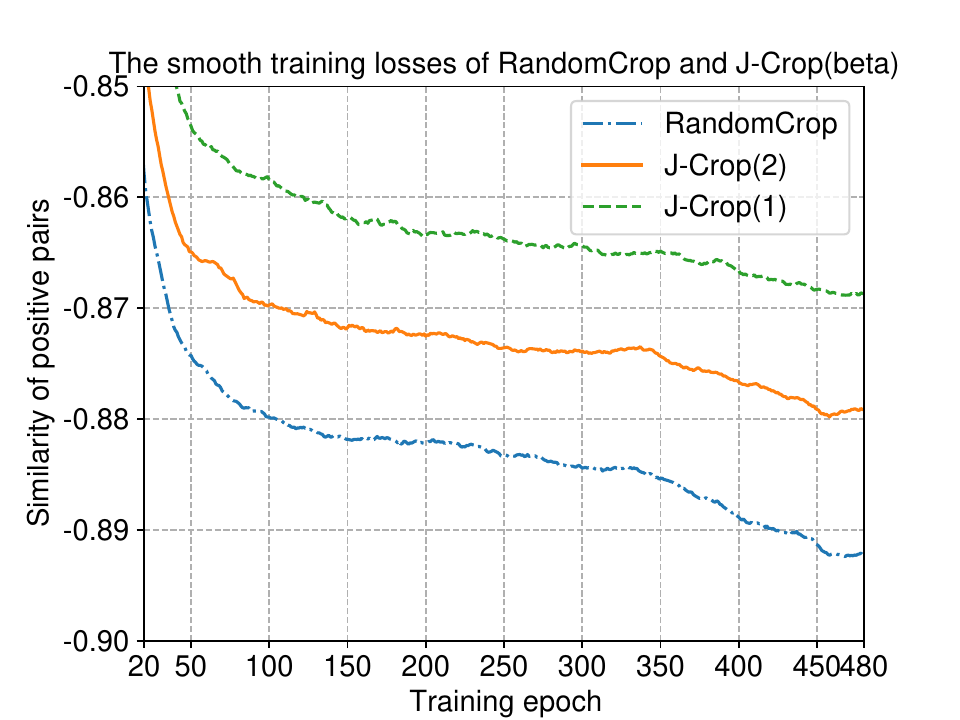}
    \caption{Training losses during SimSiam training on Tiny-ImageNet with samples generated by J-Crop$(\beta)$. We smooth the losses using a sliding window with a window size of 20. Our JointCrop creates positive pairs that are more challenging than those generated by RandomCrop.}
    \label{fig:iter_k_q_simsiam}
\end{figure}

We used the samples generated by J-Crop$(\beta)$ to train SimSiam from scratch for 500 epochs on the Tiny-ImageNet dataset, using ResNet-18 as the backbone. The training losses for different $\beta$ in J-Crop$(\beta)$ are shown in \cref{fig:iter_k_q_simsiam}. A larger loss indicates more challenging samples, and our JointCrop method indeed provides more challenging samples compared to RandomCrop. The smaller the $\beta$, the more challenging the samples are.

\subsection{JointBlur}
\label{sec:other_augmentations}

As analyzed in \cref{fig:fig1e,fig:fig1f}, we aim to develop a data augmentation method, JointBlur, that is more challenging than GaussianBlur. Chen \etal \cite{simclr} initially found that using GaussianBlur improves the classification accuracy on ImageNet-1K with ResNet-50 trained for 100 epochs. Subsequent contrastive learning methods have adopted these settings, where the image is blurred with a $50\%$ probability using a Gaussian kernel with a standard deviation $\sigma\sim\mathcal{U}[0.1, 2.0]$, and the kernel size is set to $10\%$ of the image's height and width.

GaussianBlur acts as a low-pass filter that alters the texture information of an image to varying degrees depending on its standard deviation. Our goal is to create more challenging samples for contrastive learning by distinguishing the blur levels more clearly between the two views.

Similar to JointCrop, JointBlur controls the ratio of ${\sigma_1}$ and ${\sigma_2}$, such that ${\sigma_1}, {\sigma_2}\sim\text{JC}(\beta)$ as shown in \cref{alg:jc,fig:pdf_jointcrop}. In this context, we replace $s_\text{min}$ and $s_\text{max}$ in \cref{alg:jc} with $\sigma_{min}$ and $\sigma_{max}$, set to default values of 0.1 and 2.0, respectively, which are the lower and upper bounds for the standard deviation of the Gaussian kernel.

%------------------------------------------------------------------------
\subsection{A Unified JointAugmentation Framework}
\label{sec:jointaugmentation}

In this section, we aim to abstract the common concept found in JointCrop and JointBlur into a unified framework called JointAugmentation. This approach will facilitate the application of this idea to other data augmentation methods.

In previous studies \cite{mocov1,simclr,byol,simsiam,swav}, positive sample pairs were randomly sampled with data augmentation parameters ${\bm{t_1}}$ and ${\bm{t_2}}$ independently, meaning $f({\bm{t_2}}|{\bm{t_1}})=f({\bm{t_2}})$, where $f$ represents the probability density function. In other words, when generating $v_{I,2}$, previous contrastive learning work did not utilize the known ${\bm{t_1}}$ and instead randomly and independently sampled to obtain ${\bm{t_2}}$, as described in \cref{eq:ramdom_positive_paris}. This approach could result in positive pairs that are not sufficiently challenging.

As suggested by Zhu \etal, more challenging samples may help CL learn better representations \cite{zhu2021improving}. To efficiently generate more challenging samples, JointAugmentation generates positive samples as described in \cref{eq:joint_positive_pairs}, where $G$ represents the cumulative distribution function. 

\begin{equation}
    \begin{aligned}
        v_{I,1}=\mathcal{T}\left(I;{\bm{t_1}}\right)&, v_{I,2}=\mathcal{T}\left(I;{\bm{t_2}}\right)\\
        {\bm{t_1},\bm{t_2}}\sim{G(\bm{t_1}, \bm{t_2})}&, f(\bm{t_2}|\bm{t_1})\neq f(\bm{t_2})
    \end{aligned}
    \label{eq:joint_positive_pairs}
\end{equation}

The main difference between previous methods and JointAugmentation lies in whether the known information about $v_{I,1}$ (\ie ${\bm{t_1}}$) is considered when generating $v_{I,2}$ (\ie sampling ${\bm{t_2}}$). In previous methods, the joint distribution of ${\bm{t_1}}$ and ${\bm{t_2}}$ is assumed to be the product of their marginal distributions, \ie $f({\bm{t_1}},{\bm{t_2}})=f({\bm{t_1}})\cdot f({\bm{t_2}})$. In contrast, JointAugmentation does not follow this assumption, as, $f({\bm{t_1}},{\bm{t_2}})\neq f({\bm{t_1}})\cdot f({\bm{t_2}})$. This interdependence in sampling ${\bm{t_2}}$ based on ${\bm{t_1}}$ is what gives JointAugmentation its name.

In JointCrop, ${\bm{t}}$ comprises four parameters: $i$, $j$, $w$, and $h$. We indirectly control the joint distribution $G(\bm{t_1}, \bm{t_2})$ by managing the area ratio ${s_r} = \frac{{s_2}}{{s_1}} = \frac{{h_2}{w_2}}{{h_1}{w_1}}$. For JointBlur, ${\bm{t}}$ represents the GaussianBlur kernel, and we similarly manage the joint distribution $G(\bm{t_1}, \bm{t_2})$ by controlling the ratio of the GaussianBlur kernel $\frac{{\sigma_2}}{{\sigma_1}}$. For other data augmentation methods, we begin with the augmentation parameter ${\bm{t}}$ and indirectly control the joint distribution of the two augmentation parameters $G(\bm{t_1}, \bm{t_2})$ by regulating certain aspects related to difficulty.

\section{Experiments}
\label{sec:experiments}

\subsection{Experiment Settings}
\label{sec:experiments-settings}
\textbf{Datasets}. We conduct experiments on the STL-10 \cite{stl10}, Tiny-ImageNet, and ImageNet-1K \cite{imagenet}. We also evaluate the performance of downstream tasks on PASCAL VOC \cite{pascalvoc} and COCO \cite{coco}. For specific ablation experiments, we utilize the ImageNet-100, which is created by randomly selecting 100 classes from ImageNet-1K.

\noindent
\textbf{Models}. We perform experiments on several recent and popular CL methods, including SimCLR \cite{simclr}, MoCo \cite{mocov1, mocov2, mocov3}, SimSiam \cite{simsiam}, BYOL \cite{byol}, and Dino \cite{dino}. For all these methods, we use ResNet as the backbone. The experimental setup remains consistent with the baseline, except for the modifications introduced by our JointCrop and JointBlur methods.

\noindent
\textbf{Strategies for Selecting Hyperparameters $\beta$}. As a plug-and-play method, we aim to use a common hyperparameter $\beta$ rather than having different hyperparameter choices for various situations. Therefore, unless specifically mentioned for ablation experiments, we set $\beta=0$ for both JointCrop and JointBlur. While carefully adjusting hyperparameters could potentially enhance performance, this is beyond the scope of the current work.

\noindent
\textbf{Pretraining on Small Datasets}. On small datasets such as STL-10 and Tiny-ImageNet, we use ResNet-18 as the backbone and train for 500 epochs with a batch size of 512 and a cosine annealing learning rate of 0.5.

\noindent
\textbf{Pretraining on ImageNet-1K.} On ImageNet-1K, we use ResNet-50 as the backbone. For each baseline, we adhered to the experimental setup as described in the original papers.

\noindent
\textbf{Linear Evaluation.} We evaluate the performance of the pre-trained model using linear evaluation. Specifically, we assess the top-1 accuracy of a linear classifier on the validation sets to gauge the model's performance. For linear classification on smaller datasets, such as STL-10 and Tiny-ImageNet, we use a small initial learning rate of 10. For linear classification on ImageNet-1K, we adhere to the evaluation settings outlined in the original papers.

%------------------------------------------------------------------------

\subsection{Results of JointCrop and JointBlur}
\label{sec:results_crop_blur}

\noindent
\textbf{Results on ImageNet-1K}. We applied JointBlur and JointCrop separately to the data augmentation process for ImageNet, and the results are presented in \cref{tab:JointAugImageNet}. The consistent and non-trivial improvements observed demonstrate the effectiveness of our JointCrop and JointBlur methods. We also provide results for MoCo v3 pre-trained for 300 epochs. Even under long-term training, our JointCrop continues to show significant improvements.

\begin{table}[h]
  \centering
  \resizebox{\linewidth}{!}{
  \begin{tabular}{c|c|c|c|c|c}
  \toprule
\multirow{2}{*}{Model} & Batch & \multirow{2}{*}{Epoch} & \multirow{2}{*}{Baseline} & Baseline & Baseline \\
& Size & & & +J-Crop(0) & +J-Blur(0) \\
\midrule
SimCLR & 512 & 100 & 60.7 & \textbf{62.16}\textcolor{ForestGreen}{\scriptsize (+1.46)} & 61.40\textcolor{ForestGreen}{\scriptsize (+0.70)}\\
\midrule
SimSiam & 256 & 100 & 68.1 & \textbf{68.51}\textcolor{ForestGreen}{\scriptsize (+0.41)} & 68.31\textcolor{ForestGreen}{\scriptsize (+0.21)} \\
\midrule
MoCo v1 & 256 & 200 & 60.6 & \textbf{63.29}\textcolor{ForestGreen}{\scriptsize (+2.69)} & 62.87\textcolor{ForestGreen}{\scriptsize (+2.27)} \\
\midrule
MoCo v2 & 256 & 200  & 67.5 & 67.70\textcolor{ForestGreen}{\scriptsize (+0.20)} & \textbf{67.87}\textcolor{ForestGreen}{\scriptsize (+0.37)} \\
\midrule
\multirow{2}{*}{MoCo v3} & \multirow{2}{*}{4096} & 100 & 68.9 & \textbf{69.47}\textcolor{ForestGreen}{\scriptsize (+0.57)} & -  \\
 & & 300 & 72.8 & \textbf{73.23}\textcolor{ForestGreen}{\scriptsize (+0.43)} & - \\
\bottomrule
\end{tabular}
}
\caption{Linear classification results of JointCrop and JointBlur on ImageNet-1K. All baseline results were sourced from their papers. Since MoCo v3 uses Gaussian Blur only with 10\% probability when generating $v_{I,2}$, JointBlur does not have a significant impact on MoCo v3, as marked by ``-''.}
\label{tab:JointAugImageNet}
\end{table}

\noindent
\textbf{Results on Small Datasets}. The results of the baselines and JointCrop on small datasets are shown in \cref{tab:JointCropSmall}. 
The results show that our proposed JointCrop can consistently improve the performance on small datasets. 
Due to the low resolution of images in smaller datasets, their original data augmentations do not include GaussianBlur. Consequently, using JointBlur on these small datasets is not suitable.

\begin{table}[h]
  \centering
  \resizebox{\linewidth}{!}{
  \begin{tabular}{c|c|c|c|c|c}
        \toprule
        Dataset & Method & SimCLR & BYOL & MoCo v2 & SimSiam \\
        \midrule
        \multirow{2}{*}{STL-10} & Baseline & 89.40 & 91.71 & 88.11 & 88.74\\
        & +J-Crop(0) & \textbf{90.20}\textcolor{ForestGreen}{\scriptsize (+0.80)} & \textbf{92.28}\textcolor{ForestGreen}{\scriptsize (+0.57)} & \textbf{89.78}\textcolor{ForestGreen}{\scriptsize (+1.67)} & \textbf{89.08}\textcolor{ForestGreen}{\scriptsize (+0.34)} \\
        \midrule
        \multirow{2}{*}{Tiny-IN} & Baseline & 45.25 & 48.91 & 46.07 & 44.17 \\
        & +J-Crop(0) & \textbf{47.53}\textcolor{ForestGreen}{\scriptsize (+2.28)} & \textbf{49.91}\textcolor{ForestGreen}{\scriptsize (+1.00)} & \textbf{48.45}\textcolor{ForestGreen}{\scriptsize (+2.38)} & \textbf{45.73}\textcolor{ForestGreen}{\scriptsize (+1.56)}\\
        \bottomrule
    \end{tabular}
    }
    \caption{Linear classification results on small datasets. Our JointCrop consistently provides non-trivial improvements.}
  \label{tab:JointCropSmall}
\end{table}

%------------------------------------------------------------------------
\subsection{Results of Downstream Tasks}
\label{sec:downstream_tasks}

\textbf{Object Detection on PASCAL VOC.}
Our experimental settings are the same as MoCo v1, that is, the detector is Faster R-CNN \cite{fasterrcnn} with a backbone of R50-C4 \cite{maskrcnn} with 200 epochs of pre-training. We fine-tune the pretrained model with all layers end-to-end for 24K iterations using the detectron2 codebase \cite{detectron2} on the \texttt{trainval2007+2012} split and evaluate on \texttt{test2007}.
Our method achieves improvements of $+0.8\text{AP}$ and $+0.2\text{AP}$ over MoCo v1 and MoCo v2 baselines, as shown in \cref{tab:downstream_voc}.

\noindent
\textbf{Object Detection and Instance Segmentation on COCO.}
We use Mask R-CNN \cite{maskrcnn} with a R50-C4 backbone for object detection and instance segmentation on COCO. We fine-tune all layers end-to-end for 90K iterations, that is, $1\times$ schedule on \texttt{train2017} and evaluate on \texttt{val2017}. As a result, our proposed JointCrop achieves improvements  of $+0.4\text{AP}$ and $+0.2\text{AP}$ compared with MoCo v1 and MoCo v2 baselines, as shown in \cref{tab:downstream_coco}.

\begin{table}[t]
  \centering
    \begin{tabular}{c|ccc}
    \toprule
        Method & AP       & $\text{AP}_{50}$      & $\text{AP}_{75}$  \\
        \midrule
        MoCo v1 Baseline &   55.9       &     81.5       &     62.6       \\
        MoCo v1+J-Crop(0)               &     \textbf{56.7}     &   \textbf{81.9}         &      \textbf{63.3}    \\
        \midrule
        MoCo v2 Baseline &      57.0     &        82.4     &        63.6       \\
        MoCo v2+J-Crop(0)        &        \textbf{57.2}      &     \textbf{82.5}        &     \textbf{63.9}    \\
        \bottomrule
    \end{tabular}
    \caption{Results of transferring to PASCAL VOC.}
    \label{tab:downstream_voc}
\end{table}

\begin{table}[t]
  \centering
  \resizebox{\linewidth}{!}{
    \begin{tabular}{c|ccc|ccc}
    \toprule
        \multirow{2}{*}{Method} &  \multicolumn{3}{c|}{COCO instance seg.} & \multicolumn{3}{c}{COCO detection} \\
                                   & $\text{AP}^{mk}$     &      $\text{AP}^{mk}_{50}$       &      $\text{AP}^{mk}_{75}$       &    $\text{AP}^{bb}$        &      $\text{AP}^{bb}_{50}$     &     $\text{AP}^{bb}_{75}$      \\
        \midrule
        MoCo v1 Baseline &    33.6    &        54.8     &      35.6       &     38.5     &     58.3      &     41.6     \\
        MoCo v1+J-Crop(0)  &   \textbf{34.0}        &     \textbf{55.3}      &        \textbf{36.1}    &     \textbf{38.9}       &     \textbf{58.5}      &   \textbf{42.2}        \\
        \midrule
        MoCo v2 Baseline &     34.0       &       55.4      &      36.0       &       39.0      &   58.5       &        42.3    \\
        MoCo v2+J-Crop(0)        &     \textbf{34.2}        &      \textbf{55.5}       &           \textbf{36.4}  &        \textbf{39.2}    &      \textbf{58.8}     &     \textbf{42.6}     \\
        \bottomrule
    \end{tabular}
    }
    \caption{Results of transferring to COCO detection and segmentation. We fine-tune 90K iterations, \ie, $1\times$ schedule on \texttt{train2017} and evaluate on \texttt{val2017}.}
    \label{tab:downstream_coco}
\end{table}

%------------------------------------------------------------------------

\subsection{Ablation Studies of Hyper-Parameter}
\label{sec:hyper-parameters}

Our proposed JointCrop method controls the statistical difficulty of positive pairs by adjusting $\beta$ in $\text{J-Crop}(\beta)$. We investigate the influence of $\beta$ in $\text{J-Crop}(\beta)$ on Tiny-ImageNet, as shown in \cref{tab:sigma_in_jointcrop}. A smaller $\beta$ results in more challenging samples. 
\Cref{tab:sigma_in_jointcrop} reveals an optimal point, $\beta_{op}$. Values of $\beta>\beta_{op}$ lead to oversimplified samples, while values less than $\beta<\beta_{op}$ produce samples that are too challenging, hindering the learning of better representations. As a plug-and-play framework, JointCrop is designed to be insensitive to hyperparameters and should not require a complex hyperparameter selection strategy. Across a broad range of $\beta\in[-2, 2]$, JointCrop consistently enhances the baseline. The best results, corresponding to specific $\beta$ values, are highlighted in bold. Our generic hyperparameter, $\beta=0$, is marked with yellow cells.

\begin{table}[h]
    \centering
    \resizebox{\linewidth}{!}{
    \begin{tabular}{c|c|c|c|c}
        \toprule
        Method & SimCLR & BYOL & MoCo v2 & SimSiam \\
        \midrule
        Baseline & \cellcolor{gray!30}45.25 & \cellcolor{gray!30}48.91 & \cellcolor{gray!30}46.07 & \cellcolor{gray!30}44.17\\
        \midrule
        +J-Crop(2) & 46.69 & 50.19 & 46.65 & 45.45 \\
        +J-Crop(1) & 47.77 & 49.98 & 47.81 & 45.10 \\
        +J-Crop(0) & \cellcolor{yellow!60}47.53 & \cellcolor{yellow!60}49.91 & \cellcolor{yellow!60}48.45 & \cellcolor{yellow!60}45.73\\
        +J-Crop(-1) & 47.29 & 49.54 & 48.66 & \textbf{45.91} \\
        +J-Crop(-2) & \textbf{47.95} & \textbf{51.02} & \textbf{48.78} & 45.62 \\
        \bottomrule
    \end{tabular}
    }
    \caption{Linear evaluation accuracy on Tiny-ImageNet \wrt J-Crop($\beta$). The yellow cells indicate the common hyperparameters $\beta=0$. Within the range of $\beta\in[-2, 2]$, JointCrop consistently improves the baseline, which demonstrates the plug-and-play JointCrop is not sensitive to hyperparameters.}
    \label{tab:sigma_in_jointcrop}
\end{table}

\subsection{Generalization to Other Datasets and Augmentation Methods}
\label{sec:generalization}

\noindent
\textbf{Pre-training on Non-Object-Centered and Multi-Object Datasets}. Our pre-training primarily utilizes the ImageNet-1K dataset, which is characterized by being single-object and object-centered. However, existing pre-training approaches often use images sourced from the web, which do not necessarily share these properties. This raises the question of whether our approach can be effectively applied to a broader range of contrastive learning pre-training scenarios that involve non-object-centered and multi-object images. To explore this, we utilized the COCO dataset, which is non-object-centered and contains multiple objects, for pre-training. We then fine-tuned the model on ImageNet-100. As shown in \Cref{tab:rebuttal_coco}, even when pre-training on non-object-centered and multi-object datasets, our JointCrop method demonstrates a significant improvement over the baseline.

\begin{table}[h]
    \centering
    \resizebox{\linewidth}{!}{
    \begin{tabular}{c|c|c||c|c}
    \toprule
        Method & MoCo v1 & +J-Crop(0) & MoCo v2 & +J-Crop(0) \\
        \midrule
        Accuracy & 59.62 & 61.92 & 56.12 & 57.92 \\
        \bottomrule
    \end{tabular}
    }
    \caption{Results of IN-100 when pre-training on COCO.}
    \label{tab:rebuttal_coco}
\end{table}

\noindent
\textbf{Generalizing JointAugmentation to other data augmentations}. 
In \cref{sec:jointaugmentation}, we abstract the ideas of JointCrop and JointBlur into a unified framework, JointAugmentation, which we attempt to generalize to other popular data augmentation approaches. ColorJitter performs random distortion of the image's color, with brightness $b$ and contrast $c$ being two key parameters. For example, brightness is sampled as $b\sim\mathcal{U}[1-b_f, 1+b_f]$, where $b_f\in[0,1]$ and defaults to $b_f=0.4$. The baseline ColorJitter independently performs the same operation twice to obtain ${b_1, c_1}$ and ${b_2, c_2}$. Similar to JointCrop, JointColor controls the joint distribution of $b_1$ and $b_2$ ($c_1$ and $c_2$) through the ratios $b_1/b_2$ ($c_1/c_2$). The non-trivial improvements presented in \cref{tab:rebuttal_jointcolor} further demonstrate the generalizability of our JointAugmentation framework to other data augmentation methods.

\begin{table}[!h]
    \centering
    % \resizebox{\linewidth}{!}{
    \begin{tabular}{c|c|c|c}
    \toprule
        Method & Baseline & JointColor-$b$ & JointColor-$c$\\
        \midrule
        MoCo v1 & 63.18 & 63.80 & 63.68\\
        \bottomrule
    \end{tabular}
    % }
    \caption{Results of JointColor on ImageNet-100.}
    \label{tab:rebuttal_jointcolor}
\end{table}

\subsection{Potential Combinations With Other Methods}
\label{sec:combinations}

\noindent
\textbf{In-depth analysis of JointCrop and MultiCrop}. JointCrop and MultiCrop have different motivations and methods. They can be used in combination leading to better representations, as in \cref{sec:multicrop_jointcrop}.

\noindent
\textbf{Combined use of JointCrop and JointBlur}. The combination of JointCrop and JointBlur requires more fine-grained considerations, otherwise it may produce overly difficult samples, as in \cref{sec:jointcrop_jointblur}.

\noindent
\textbf{Combined use of JointCrop and InfoMin}. The combination of JointCrop with InfoMin further improves the InfoMin baseline from 67.4 to 67.81.

\noindent
\textbf{In-depth analysis of JointCrop and ContrastiveCrop}. We provide an in-depth analysis for ContrastiveCrop and our JointCrop in \cref{sec:jointcrop_ccrop}, and try to combine the two.

\noindent
\textbf{Analysis of positive pair distances}. JointCrop actually controls not only the area ratio, but also implicitly controls the distance between positive samples, as in \cref{sec:jointcrop_distance}.

\subsection{Computational Complexity Analysis}
\label{sec:complexity}

\noindent
We train 5 epochs for the MoCo v1 baseline, JointCrop, and JointBlur on and calculated the average running time. \Cref{tab:rebuttal_training_time} illustrates our JointCrop and JointBlur introduce little additional computational complexity. This does not indicate our method is faster, since the time differences are minor.

\begin{table}[!h]
    \centering
    \resizebox{\linewidth}{!}{
    \begin{tabular}{c|c|c|c}
    \toprule
        Method &  MoCo v1 & +JointCrop & +JointBlur\\
        \midrule
        Time (seconds) &  788.2{\scriptsize($\pm$2.2)} & 782.8{\scriptsize($\pm$1.8)} & 785.8{\scriptsize($\pm$1.8)} \\
        \bottomrule
    \end{tabular}
    }
    \caption{Training time of baseline and our methods.}
    \label{tab:rebuttal_training_time}
\end{table}

%------------------------------------------------------------------------

\section{Conclusion}
\label{sec:conclusion}

In this work, we propose JointCrop and JointBlur, which explore the correlation between two augmentations of positive pairs and generates more challenging positive pairs by controlling the joint distribution of these augmentation parameters. We also integrated both approaches into a unified framework called JointAugmentation, paving the way for applying this concept to other forms of data enhancement. The effectiveness of our method has been demonstrated across multiple popular contrastive learning methods. 
We hope our work will inspire further research on data augmentations in contrastive learning. 

Our limitations and future work are in \cref{sec:limitation}.

\section*{Acknowledgements}

The authors would like to thank Xuefei Ning, Cheng Yu, Youze Xue and Yu Shang for their help in revising the paper.

This work was supported by the National Natural Science Foundation of China (62376024, 62325405), the Young Elite Scientists Sponsorship Program by CAST (2023QNRC001) and Beijing National Research Center for Information Science and Technology (BNRist, BNR2024TD03001).

\bibliography{aaai25}

\clearpage

\appendix

\section{Detailed steps of the JointCrop}
\label{sec:algorithm}
For a clearer understanding of JointCrop, we provide the detailed steps in \cref{alg:jc}.

\begin{algorithm}[h]
    \SetKwInput{KwHyperparameter}{Hyperparameter}
    \caption{The steps of J-Crop$(\beta)$ different from RandomCrop.}
    \label{alg:jc}
    \KwHyperparameter{the area range $[s_\text{min},s_\text{max}]$, the $\beta$ controls the difficulty}
    \KwOut{Areas of a pair of positive views ${s_1}$ and ${s_2}$}
    \tcc{In RandomCrop, $s_1$ and $s_2$ are independently and identically distributed and their ratio is likely to be close to 1}
    ${s_1} \sim \mathcal{U}[s_\text{min},s_\text{max}]$\;
    ${s_2} \sim \mathcal{U}[s_\text{min},s_\text{max}]$\;
    \Return{${s_1},{s_2}$}\;
    \tcc{JointCrop controls the joint distribution of ${s_1}$ and ${s_2}$ by controlling the area ratio $s_r=s_2/s_1$}
    \setcounter{AlgoLine}{0} % 重新编号
    $s_b=\log\dfrac{s_\text{max}}{s_\text{min}}$\;
    \uIf{$\beta>0$}{
        $\log{s_r}\sim \mathcal{N}_T(0,\frac{1}{\beta}s_b,-s_b,s_b)$\;
    }
    \uElseIf{$\beta=0$}{
        $\log{s_r}\sim \mathcal{U}[-s_b,s_b]$\;
    }
    \ElseIf{$\beta<0$}{
        $\log{s_r}\sim \mathcal{N}_{T}(0,-\frac{1}{\beta}s_b,-s_b,s_b)$\;
        \eIf{$\log{s_r}<0$}
        {$\log{s_r} = -s_b - \log{s_r}$\;}
        {$\log{s_r} = s_b - \log{s_r}$\;}
    }
    ${s_1} \sim \mathcal{U} \left[\max\left(s_\text{min},\frac{s_\text{min}}{{s_r}}\right), \min\left(\frac{s_\text{max}}{{s_r}},s_\text{max}\right)\right]$\;
    ${s_2} = {s_1} \times {s_r}$\;
    \Return{${s_1},{s_2}$}\;
    \tcc{With ${s_1}$ and ${s_2}$ known, we can use the same steps as RandomCrop to compute ${h_1}$, ${w_1}$, ${i_1}$, ${j_1}$ and then get a pair of positive views}
\end{algorithm}

\section{Code of Our Method}
Different code frameworks are employed for various methods, as illustrated in \cref{tab:code_base}. 
The official code is directly cloned, and modifications are made to the data augmentation components of these codes. 
For instance, in replacing RandomCrop with JointCrop, it is necessary to overload the RandomCrop class `torchvision.transforms.RandomResizedCrop' and amalgamate the cropping process for a pair of positive samples into a singular class or function to achieve a `joint' operation. Only this segment of the code requires modification, while the remainder of the code is maintained as per the baseline code utilized.

The code for pre-training with JointCrop and performing linear fine-tuning on MoCo v3 has been included in the supplementary material. Detailed instructions for setting up the environment and executing the code are provided in the README. We used different codebases for various baselines, as outlined in \cref{tab:code_base}.

\begin{table}[h]
    \centering
    \begin{tabular}{cc}
        \toprule
        {Method}      &  {Code Link}          \\
        \midrule
        MoCo v1  &  https://github.com/facebookresearch/moco \\
        MoCo v2  &  https://github.com/facebookresearch/moco \\
        MoCo v3  &  https://github.com/facebookresearch/moco-v3 \\
        SimSiam  &  https://github.com/open-mmlab/mmselfsup \\
        BYOL  &   https://github.com/open-mmlab/mmselfsup \\
        SimCLR  &  https://github.com/open-mmlab/mmselfsup \\
        Dino & https://github.com/facebookresearch/dino \\
        % downstream tasks   &  https://github.com/facebookresearch/moco/tree/main/detection \\
        \bottomrule
    \end{tabular}
    \caption{The code base we use for various methods.}
    \label{tab:code_base}
\end{table}

\section{Limitation and Future Work}
\label{sec:limitation}
The limitations of this paper and the directions for our future work are summarized as follows.
\subsection{Limitation}
\begin{itemize}
    \item Our approach necessitates pre-training from scratch, consuming significant energy, time, and computational resources. Nonetheless, we intend to make the pre-trained weights publicly accessible, enabling others to fine-tune their downstream tasks accordingly. 
    \item Instance Discrimination (ID) and Masked Image Modeling (MIM) represent prominent paradigms in self-supervised learning (SSL). Our methodology is exclusively adaptable to ID (contrastive learning) and is not applicable to MIM (generative SSL, such as MAE).
\end{itemize}
\subsection{Future work}
\begin{itemize}
    \item We aim to achieve outcomes from training over extended epochs. However, such an endeavor demands considerable time and computational resources. The most extensive pre-training demonstrated herein is the outcome of 300 epochs of MoCo v3 pre-training.
    \item We plan to investigate the integration of various JointAugmentation techniques to generate challenging pairs of positive samples, thereby enhancing the model's feature representation capabilities.
\end{itemize}

\section{Analysis for JointCrop and MultiCrop}
\label{sec:multicrop_jointcrop}

Multi-Crop enables positive pairs to have both a global view and a local view by manually specifying the area size during cropping and using more than two views. For example, a typical setup includes two global views with an area ranging from 40\% to 100\% of the original image and four local views with an area between 5\% and 40\% of the image. The key distinction between Multi-Crop and our JointCrop is that: (1) In Multi-Crop, the data augmentations for positive pairs are independent, whereas in JointCrop, they are not. (2) JointCrop incurs no additional computational cost, whereas Multi-Crop takes approximately 1.35 times longer than the baseline. (3) The underlying concepts of our JointCrop can be extended to other augmentations, such as JointBlur and JointColor, among others. In contrast, Multi-Crop lacks this kind of extensibility. (4) JointCrop can be further integrated with Multi-Crop to enhance its capabilities. We tested the combination of JointCrop and Multi-Crop. Specifically, we applied JointCrop to both the global and local pairs of Dino \cite{dino} with the same Multi-Crop configuration of 2x160+4x96, and then trained and fine-tuned on ImageNet-1K. As shown in \Cref{tab:rebuttal_multicrop}, our JointCrop can be effectively combined with Multi-Crop to further enhance the performance of contrastive learning. 

\begin{table}[!h]
    \centering
    % \resizebox{\linewidth}{!}{
    \begin{tabular}{c|c|c}
    \toprule
        Model & Multi-Crop & Multi-Crop+JointCrop \\
        \midrule
        \multirow{2}{*}{Dino} & \multirow{2}{*}{66.36} & 66.55 (global) \\
        & & 66.58 (local) \\
        \bottomrule
    \end{tabular}
    % }
    \caption{Results of the Combination of Multi-Crop and JointCrop. As these baseline were not available in their original papers, we have reproduced them for our analysis.}
    \label{tab:rebuttal_multicrop}
\end{table}

\section{Analysis for Distance between Positive Pairs in JointCrop}
\label{sec:jointcrop_distance}

\begin{figure}[h]
    \centering
    \begin{subfigure}[b]{\linewidth}
        \includegraphics[width=\textwidth]{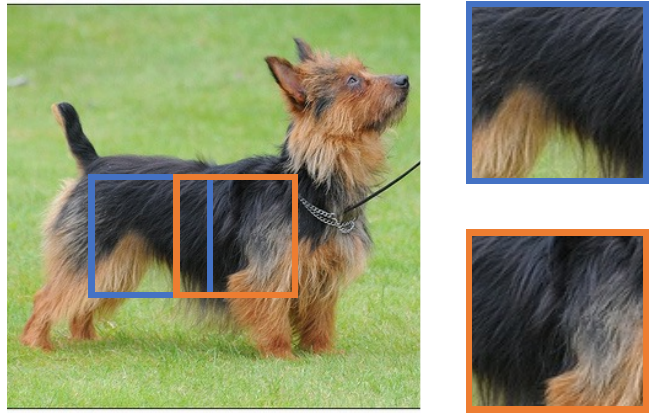}
        \caption{Close views may be similar.}
        \label{fig:8a}
    \end{subfigure}
    \hfill
    \begin{subfigure}[b]{\linewidth}
        \includegraphics[width=\textwidth]{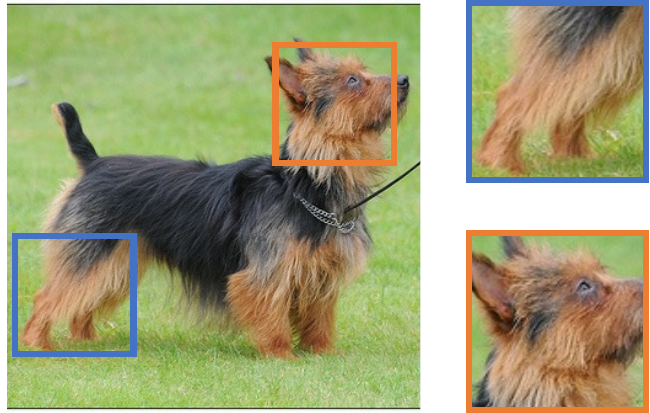}
        \caption{Distant views may be challenging.}
        \label{fig:8b}
    \end{subfigure}
    \caption{Distance between views may affect difficulty.}
    \label{fig:figure8}
\end{figure}

\begin{figure}[h]
    \centering
    \includegraphics[width=\linewidth]{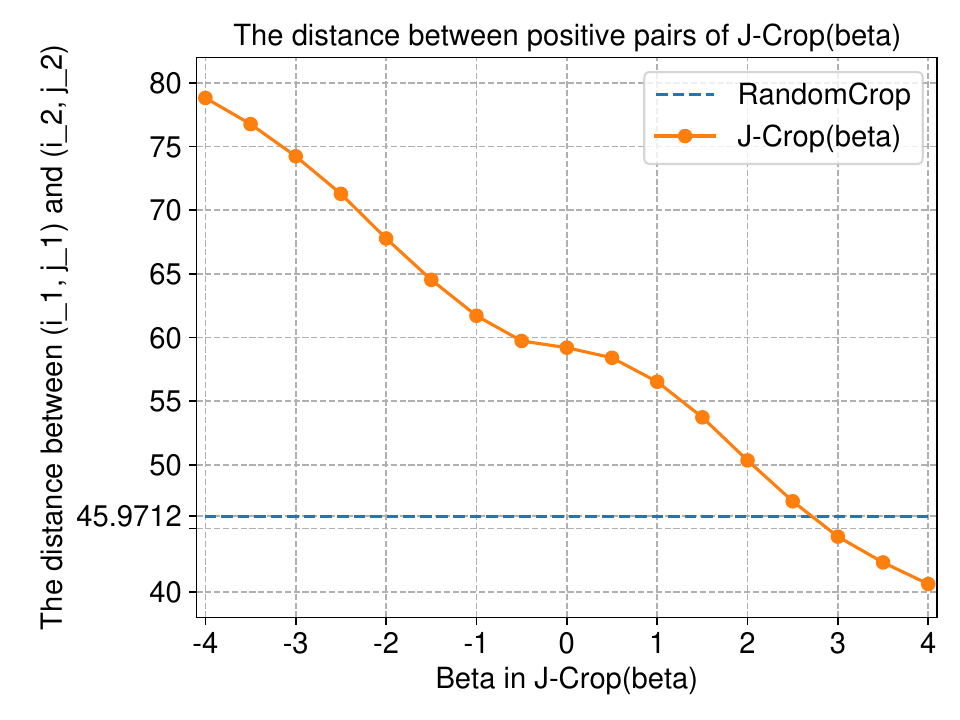}
    \caption{The distance between pairs also affects difficulty. JointCrop not only controls the area ratios directly, but also indirectly controls the distance between positive pairs.}
    \label{fig:distance_jcrop}
\end{figure}

In JointCrop, we control ${h_1}$, ${w_1}$, and ${h_2}$, ${w_2}$ by controlling the area ratios ${s_1}$ and ${s_2}$. However, Crop requires sample locations $\left({i_1}, {j_1}\right)$ and $\left({i_2}, {j_2}\right)$ in addition to sample widths and heights. The Euclidean distance $d=\sqrt{{\left({i_1}-{i_2}\right)}^2+{\left({j_1}-{j_2}\right)}^2}$ of the positional coordinates between a positive pair also affects its difficulty. A pair of positive samples that are close together may have similar features, \eg, the two parts of a dog's body shown in \cref{fig:8a}. This pair is itself very similar and simple for CL. Whereas a pair of positive samples that are farther away may have dissimilar features but have the same semantic information (they are on one image, after all), \eg, the dog's head and dog's leg in \cref{fig:8b}. This pair is more challenging and may help models learn better. 

Can we add control over distance in JointCrop to get more challenging samples? 
This is certainly possible, but let's first note that JointCrop actually implicitly controls the distance already.
In \cref{sec:preliminary} we have shown that ${i}\sim\mathcal{U}[0,\text{W}-{w}]$ and ${j}\sim\mathcal{U}[0,\text{H}-{h}]$. Compared to RandomCrop, JointCrop directly controls the area ratio ${s_2}/{s_1}$ and indirectly changes the distribution of ${s_1}$. This indirectly changes the distributions of ${h_1}$ and ${w_1}$ and then affects the distributions of ${i_1}$ and ${j_1}$. Similarly, the distributions of ${i_2}$ and ${j_2}$ are indirectly controlled. 
Therefore, JointCrop will also affect distance $d$. Direct theoretical derivation of distances is very complex. 
We repeated the experiment three times, each time taking 100,000 positive pairs from J-Crop$(\beta)$ and measuring the distance between them, as illustrated in \cref{fig:distance_jcrop}. A smaller $\beta$ does increase the Euclidean distance between pairs of positive samples.

\section{Analysis for the Combination of JointCrop and JointBlur}
\label{sec:jointcrop_jointblur}

In this paper we propose JointCrop and JointBlur, and both JointCrop and JointBlur can improve the linear evaluation accuracy over baselines under multiple datasets and on multiple CL methods. However, what if we use them together, \eg, the combination of JointCrop and JointBlur?

In fact, these two augmentation methods are in conflict with each other. 
JointCrop encourages the ratio of the area of a positive pair of samples to be farther away from 1, that is, with a higher probability of obtaining a ``large view'' and a ``small view''. The ``large'' and ``small'' here do not refer to the size of the images, as they are all resized to the same size (default to $224\times224$). They refer to the area of the views in the original image. 
While, JointBlur makes the GaussianBlur kernel of a pair of positive samples more different, that is, with a higher probability of obtaining a ``fuzzy view'' and a ``clear view''. 
If we use the two methods at the same time, we may obtain overly challenging samples, \eg, pairs of positive samples with a ``small fuzzy view'' and a ``large clear view'', which are too difficult to learn good feature representations. 

\begin{figure}[h]
  \centering
  \begin{subfigure}{0.32\linewidth}
    % \fbox{\rule{0pt}{2in} \rule{.9\linewidth}{0pt}}
    \includegraphics[width=\linewidth]{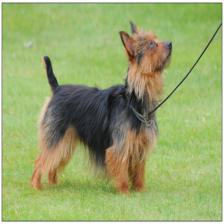}
    \caption{}
    \label{fig:ori}
  \end{subfigure}
  \hfill
  \begin{subfigure}{0.32\linewidth}
    % \fbox{\rule{0pt}{2in} \rule{.9\linewidth}{0pt}}
    \includegraphics[width=\linewidth]{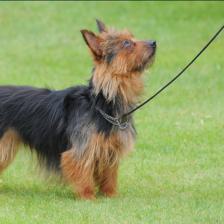}
    \caption{}
    \label{fig:large}
  \end{subfigure}
  \hfill
  \begin{subfigure}{0.32\linewidth}
    % \fbox{\rule{0pt}{2in} \rule{.9\linewidth}{0pt}}
    \includegraphics[width=\linewidth]{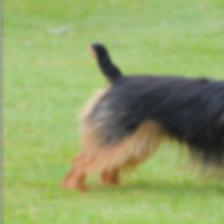}
    \caption{}
    \label{fig:small}
  \end{subfigure}
  \begin{subfigure}{0.32\linewidth}
    % \fbox{\rule{0pt}{2in} \rule{.9\linewidth}{0pt}}
    \includegraphics[width=\linewidth]{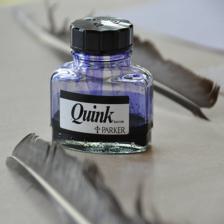}
    \caption{}
    \label{fig:ori-2}
  \end{subfigure}
  \hfill
  \begin{subfigure}{0.32\linewidth}
    % \fbox{\rule{0pt}{2in} \rule{.9\linewidth}{0pt}}
    \includegraphics[width=\linewidth]{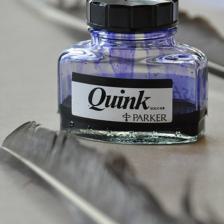}
    \caption{}
    \label{fig:large-2}
  \end{subfigure}
  \hfill
  \begin{subfigure}{0.32\linewidth}
    % \fbox{\rule{0pt}{2in} \rule{.9\linewidth}{0pt}}
    \includegraphics[width=\linewidth]{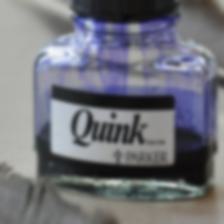}
    \caption{}
    \label{fig:small-2}
  \end{subfigure}
  \caption{Examples of pairs of positive samples with a ``small fuzzy view'' (c and f) and a ``large clear view'' (b and e). The subfigures (a) and (d) are the original images.}
  \label{fig:figure-comb}
\end{figure}

Examples of such a case are given in \Cref{fig:figure-comb}. \Cref{fig:ori} shows an image from ImageNet-1K, and \Cref{fig:large,fig:small} show a pair of positive samples obtained using JointCrop and JointBlur. \Cref{fig:large} has a larger area in the original image but uses a weaker GaussianBlur, while \Cref{fig:small} has a small area in the original image. \Cref{fig:small} is already not as high-resolution as \Cref{fig:large} when it is resized to the same size. In addition, with a stronger GaussianBlur, \Cref{fig:large,fig:small} are too difficult to learn a good feature representation. \Cref{fig:ori-2,fig:large-2,fig:small-2} are also similar examples. 

While the simultaneous application of JointCrop and JointBlur might produce samples that are excessively challenging, employing them in sequence can be beneficial. For instance, one can apply either JointCrop or JointBlur with a certain probability when generating positive samples. Our experiments, conducted using MoCo v1, demonstrate that employing J-Crop(0) or J-Blur(0) individually enhances the baseline accuracy from 57.25\% to 60.87\% and 60.58\%, respectively. Furthermore, applying either of these methods with a probability of 0.5 yields an improved result of 61.24\%.

\section{Combine JointCrop and ContrastiveCrop}
\label{sec:jointcrop_ccrop}
Our JointCrop method explicitly manages the area ratio of two cropping regions $\frac{h_2\cdot w_2}{h_1\cdot w_1}$ and implicitly influences the distance $d$ between positive pairs, in contrast to ContrastiveCrop \cite{ccrop}, which directly controls the cropped regions $i_1,j_1$ and $i_2,j_2$. ContrastiveCrop is bifurcated into two segments.
\begin{itemize}
    \item \textit{Semantic-aware Localization} restricts the cropping area through a heatmap to preclude object absence, thereby increasing the likelihood of samples appearing proximal to, particularly at the center of, the heatmap.
    \item \textit{Center-suppressed Sampling} enforces a $\beta$-distribution for coordinates $i_1,j_1$ and $i_2,j_2$ to diversify positive pairs that are overly analogous. 
\end{itemize}
 
The integration of JointCrop and Center-suppressed Sampling might yield superior outcomes, as both methods engender more challenging views. 
However, \textit{Semantic-aware Localization} by constricting the cropping area, offers less challenging views. 
The results of MoCo v1 on ImageNet-1K are presented in \cref{tab:JointCrop&ccrop}. 
Employing our JointCrop independently surpasses the MoCo v1 baseline and ContrastiveCrop, while the amalgamation of our JointCrop with \textit{Center-suppressed Sampling} (a component of ContrastiveCrop) yields enhanced results (as indicated by bolded number in \cref{tab:JointCrop&ccrop}). 
Furthermore, JointCrop is also amenable to integration with additional techniques.

\begin{table}[ht]
    \centering
    % \resizebox{\linewidth}{!}{
    \begin{tabular}{c|c}
        \toprule
        \multirow{2}{*}{Method}     &  ImageNet-1K  \\
        & Top-1 Accuracy \\
        \midrule
        MoCo v1 Baseline     &   57.25   \\
        \midrule
        MoCo v1 + & \multirow{2}{*}{58.34} \\
        ContrastiveCrop &     \\
        \midrule
        MoCo v1 + & \multirow{2}{*}{60.87} \\
        JointCrop(0) &  \\
        \midrule
        JointCrop(0) + & \multirow{2}{*}{\textbf{61.33}} \\
        \textit{Center-suppressed Sampling} &  \\
        \bottomrule
    \end{tabular}
    % }
    \caption{The combination of our proposed JointCrop and ContrastiveCrop on ImageNet-1K. The training setup is the same as MoCo v1 and ContrastiveCrop. }
    \label{tab:JointCrop&ccrop}
\end{table}

%------------------------------------------------------------------------

\section{Derivation of Probability Density Functions for Area Ratios in RandomCrop}
\label{sec:jifen}

Assume that $\widetilde{s}_1,\widetilde{s}_2\sim\mathcal{U}[s_{min},s_{max}]$, where $\mathcal{U}$ denotes a uniform distribution. Consequently, the probability density functions $g(y)$ for $\widetilde{s}_1$ and $\widetilde{s}_2$ are identical.
\begin{equation}
    g(y)=\dfrac{1}{s_{max}-s_{min}}
\end{equation}

Let the random variable $X=\dfrac{\widetilde{s}_1}{\widetilde{s}_2}$ be defined, and for $s_{min}\leq x\leq 1$, the cumulative density function $F(x)$ is specified.

\begin{align}
     F(x)&=p(X\leq x)=\int_{s_{min}/x}^{s_{max}}\int^{\widetilde{s}_2y}_{s_{min}}g^2(y)~\mathrm{d}\widetilde{s}_1\mathrm{d}\widetilde{s}_2\notag\\
         &=\frac{1}{(s_{max}-s_{min})^2}\int_{s_{min}/x}^{s_{max}}\left(\widetilde{s}_2y-s_{min}\right)\mathrm{d}\widetilde{s}_2\notag\\
         &=\frac{1}{(s_{max}-s_{min})^2}\left[\frac{x}{2}\widetilde{s}_2^2-s_{min}\widetilde{s}_2\right]_{s_{min}/x}^{s_{max}}\notag\\
         &=\frac{1}{(s_{max}-s_{min})^2}\left[\frac{s_{max}^2}{2}x-s_{max}s_{min}+\frac{s_{min}^2}{2x}\right]
     \label{test}
\end{align}
   
For $\frac{s_{min}}{s_{max}}<x<1$, the probability density function of $Y$ is derived from \cref{test}.

\begin{align}
    f(x)&=\frac{\mathrm{d}}{\mathrm{d}x}F(x)\notag\\
        &=\frac{1}{(s_{max}-s_{min})^2}\left[\frac{s_{max}^2}{2}-\frac{s_{min}^2}{2x^2}\right]\notag\\
        &=\frac{s_{max}^2x^2-s_{min}^2}{2x^2(s_{max}-s_{min})^2}
\end{align}

Similarly, $f(x)$ can be determined for $1<x\leq\frac{s_{max}}{s_{min}}$.

\begin{equation}
    f(x)=\frac{s_{max}^2-s_{min}^2x^2}{2x^2(s_{max}-s_{min})^2}
\end{equation}

In summary, the probability density function $f(x)$ for $X=\widetilde{s}_1/\widetilde{s}_2$ is presented in \cref{eq:dis_a_sign}.

\begin{equation}
      f(x)=\left\{
    \begin{aligned}
    \frac{s_{max}^2x^2-s_{min}^2}{2x^2(s_{max}-s_{min})^2} & , & \frac{s_{min}}{s_{max}}\leq x \leq 1, \\
    \frac{s_{max}^2-s_{min}^2x^2}{2x^2(s_{max}-s_{min})^2}  & , & 1< x \leq \frac{s_{max}}{s_{min}}.
    \end{aligned}
    \right.
      \label{eq:dis_a_sign}
\end{equation}

In RandomCrop, typical settings are $s_{min}=0.2$ and $s_{max}=1$. 
By substituting these values into \cref{eq:dis_a_sign}, the probability density map of $d(R\left(I_j;\widetilde{s}_{1}\right),R\left(I_j;\widetilde{s}_{2}\right))$ is obtained, as shown in \cref{eq:dis_b}.
\begin{equation}
  a\sim F(x)=\left\{
    \begin{aligned}
    \frac{25}{32}x-\frac{5}{16}+\frac{1}{32x} & , & 0.2\leq x \leq 1, \\
    -\frac{1}{32}x+\frac{21}{16}-\frac{25}{32x} & , & 1< x \leq 5.
    \end{aligned}
\right.
  \label{eq:dis_a}
\end{equation}
    \begin{equation}
      \widetilde{s}_r=\frac{\widetilde{s}_1}{\widetilde{s}_2}\sim f(x)=\left\{
    \begin{aligned}
    \frac{25}{32}-\frac{1}{32x^2} & , & 0.2\leq x \leq 1, \\
    -\frac{1}{32}+\frac{25}{32x^2} & , & 1< x \leq 5.
    \end{aligned}
    \right.
      \label{eq:dis_b}
\end{equation}

\end{document}